\documentclass[10pt,twocolumn,letterpaper]{article}

\usepackage{iccv}
\usepackage{times}
\usepackage{epsfig}
\usepackage{graphicx}
\usepackage{amsmath}
\usepackage{amssymb}
\usepackage{cite}
\usepackage{color}
\usepackage[dvipsnames]{xcolor}
\definecolor{citecolor}{HTML}{0071bc}
\usepackage{paralist, tabularx}
\usepackage{enumitem}

\usepackage{booktabs}

\usepackage{caption}

\usepackage[pagebackref=true,breaklinks=true,letterpaper=true,colorlinks,citecolor=citecolor,bookmarks=false]{hyperref}

\iccvfinalcopy 

\def\iccvPaperID{9} 

\ificcvfinal\fi

\begin{document}

\title{COSE: A Consistency-Sensitivity Metric for Saliency on Image Classification}

\author{
    Rangel Daroya$^*$ \qquad
    Aaron Sun$^*$ \qquad
    Subhransu Maji \\
    University of Massachusetts Amherst \\
    {\tt\small \{rdaroya, aaronsun\}@umass.edu, smaji@cs.umass.edu}
}

\twocolumn[{%
\renewcommand\twocolumn[1][]{#1}%
\maketitle

\begin{center}
    \centering
    \captionsetup{type=figure}
    \vspace{-6mm}
    \includegraphics[scale=0.44]{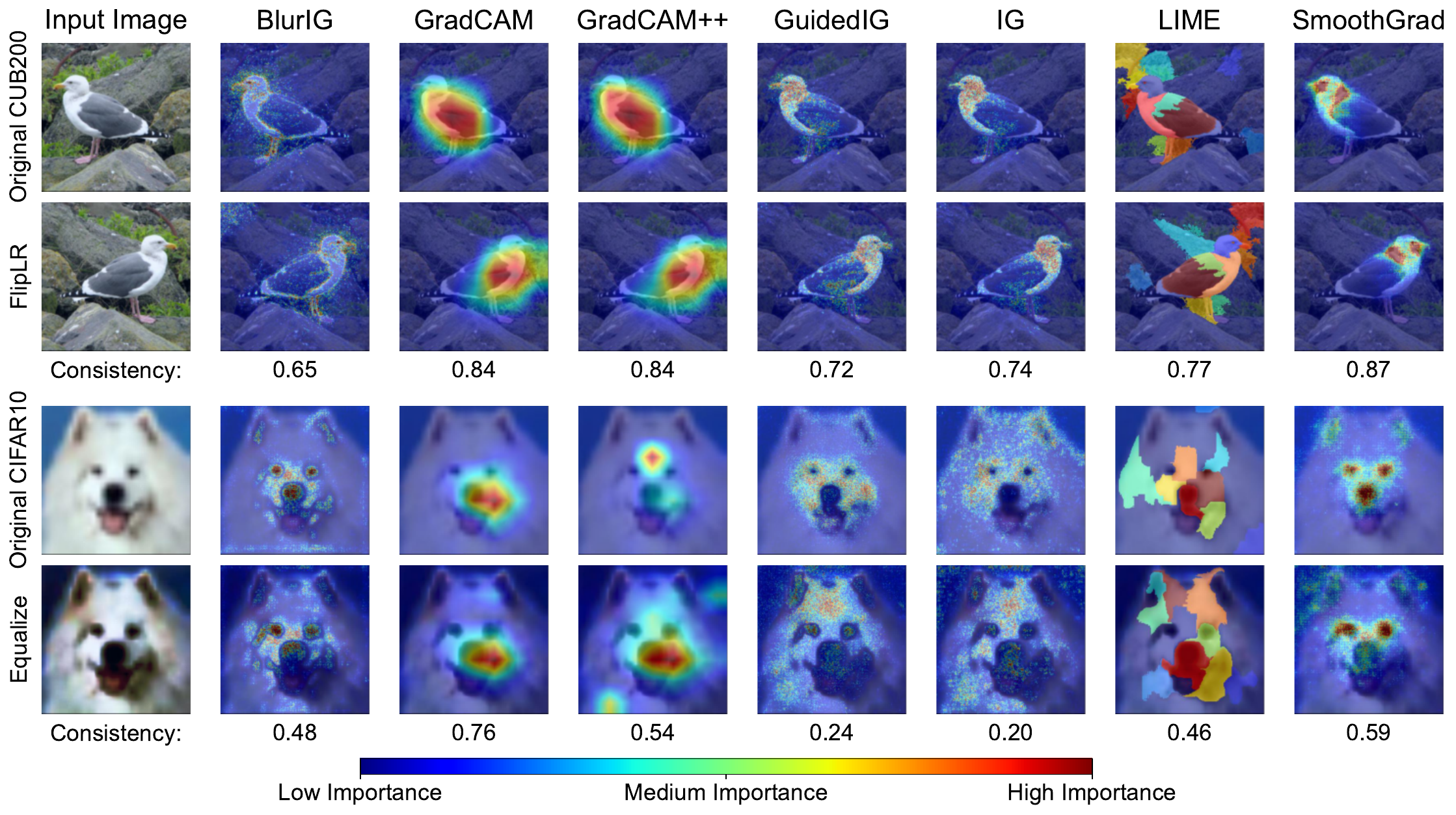}
    \captionof{figure}{\textbf{Saliency maps extracted from a ResNet50 on CUB200 and CIFAR10 images}. The colors indicate pixel importance predicted by different methods (blue=low; red=high). We show two sample data transformations: FlipLR (geometric transformation) and Equalize (photometric transformation). GradCAM \cite{selvarajuGradCAMVisualExplanations} is consistently focuses on the chest of the bird, despite the left-right flip. Even with image equalization, GradCAM emphasizes the nose of the dog. While other methods appear to do well on FlipLR, they struggle with Equalize. The proposed  \textbf{COnsistency-SEnsitivity (COSE)} metric quantifies the equivariant and invariant properties of visual model explanations using simple data augmentations.}
    \label{fig:saliency-images}
\end{center}%
}]
\ificcvfinal\thispagestyle{empty}\fi

\def\thefootnote{*}\footnotetext{Equal contribution}\def\thefootnote{\arabic{footnote}}
\begin{abstract}
   We present a set of metrics that utilize vision priors to effectively assess the performance of saliency methods on image classification tasks. 
   To understand behavior in deep learning models, many methods provide visual saliency maps emphasizing image regions that most contribute to a model prediction. 
   However, there is limited work on analyzing the reliability of saliency methods in explaining model decisions.
   We propose the metric \textbf{COnsistency-SEnsitivity (COSE)} that quantifies the equivariant and invariant properties of visual model explanations using simple data augmentations. 
   Through our metrics, we show that although saliency methods are thought to be architecture-independent, most methods could better explain transformer-based models over convolutional-based models. 
   In addition, GradCAM was found to outperform other methods in terms of COSE but was shown to have limitations such as lack of variability for fine-grained datasets. 
   The duality between consistency and sensitivity allow the analysis of saliency methods from different angles. 
   Ultimately, we find that it is important to balance these two metrics for a saliency map to faithfully show model behavior.
\end{abstract}


\section{Introduction}

Given a function $f$ operating on images $x \in \mathbb{R}^{m\times n \times c}$, a saliency map $M \in \mathbb{R}^{m\times n}$ indicates the relative importance of each pixel in the image $x$ in making the prediction $f(x)$. Saliency maps have been widely used to understand (possibly black-box) function behavior, especially with deep networks. 
They are important for humans to establish trust in predictions through transparency, and have been applied in high-stakes decisions such as medical diagnoses \cite{sundararajanAxiomaticAttributionDeep} and bias identification \cite{selvarajuGradCAMVisualExplanations}. 
Given saliency techniques are inextricably linked to human understanding, saliency maps should fulfill certain properties based on our understanding of the visual system in the world around us. 

We propose consistency and sensitivity metrics that measure two complementary properties of saliency maps. 
Consistency refers to the property that saliency maps should remain \emph{unchanged} when an input is transformed in a way that the model predictions don't change. For example, when an input is reflected or translated by a small amount, the corresponding saliency maps should also undergo the same geometric transformation, as we do not expect the class predictions to change. Similarly, when the input undergoes a photometric transformation (e.g., change in pixel intensities or blurring), we expect saliency maps to remain identical.  In short, consistency captures the degree to which saliency maps are equivariant and invariant to transformations that don't affect model predictions. Sensitivity refers to the property that saliency maps \emph{should change} when the model produces a different output. This difference in model output could be a result of changes in the model parameters (e.g., during the process of training the model) or sufficiently large changes in input. Thus, the model's explanations must change for it to produce a different output. Prior knowledge was used to identify transformations that should result in equivariance and invariance for various computer vision tasks.

While prior work has focused on evaluating consistency of saliency maps \cite{yehFidelitySensitivityExplanations2019, tomsettSanityChecksSaliency2020, zhouFeatureAttributionMethods2022, kapishnikovXRAIBetterAttributions2019}, we show that sensitivity is also a key consideration and often in conflict with consistency. We propose a combined metric called COSE defined as the harmonic mean of the consistency and sensitivity. Our work also considers natural changes to the input, and model perturbations that occur in realistic training settings.

We develop a benchmark where we evaluate several saliency methods \cite{selvarajuGradCAMVisualExplanations, chattopadhay2018grad, sundararajanAxiomaticAttributionDeep, xu2020attribution, kapishnikov2021guided, smilkov2017smoothgrad, ribeiroWhyShouldTrust2016}, deep network architectures \cite{heDeepResidualLearning2016, liu2022convnet, dosovitskiy2020vit, liu2021swin}, pre-training procedures \cite{caron2021emerging, chen2021empirical, zhou2021ibot, tian2023designing}, and evaluate these metrics on five different datasets \cite{krizhevsky2009learning, li_andreeto_ranzato_perona_2022, welinder2010caltech, helber2019eurosat, nilsback2008automated}.
We find that saliency maps generally produce more coherent explanations on transformer-based models than convolutional-based models. GradCAM also demonstrates better performance across the different metrics and across the different evaluation settings when compared to other methods. Finally, we observe common limitations among saliency methods on balancing consistency and sensitivity, and we recommend future directions for the improvement of saliency methods. In summary, our contributions include the following:
\begin{itemize}[noitemsep,topsep=0pt,parsep=0pt,partopsep=0pt,leftmargin=*]
    \item We propose the metrics \textbf{consistency}, \textbf{sensitivity}, and \textbf{COSE} to evaluate the robustness of saliency methods to input and model changes based on vision priors. 
    \item We introduce an evaluation pipeline that incorporates \textit{natural} image and model variations encountered by human end users which we open source for future research.\footnotemark{}
    \item We show the effectiveness of our proposed metrics to evaluate different model architectures (with supervised and unsupervised features) to analyze the behavior of saliency methods across different settings.
\end{itemize}
\footnotetext{The code is available at \href{https://github.com/cvl-umass/COSE}{https://github.com/cvl-umass/COSE}}

\section{Related Work}

\subsection{Saliency Maps}
Saliency explanations generally attribute importance to input features \cite{adebayoSanityChecksSaliency2018}. For images, explanations typically are represented as saliency heatmaps, in which ``important" pixels are highlighted. Most explainability methods either involve gradient and activation summation \cite{selvarajuGradCAMVisualExplanations, chattopadhay2018grad}, input perturbations \cite{ribeiroWhyShouldTrust2016}, or some combination of both \cite{sundararajanAxiomaticAttributionDeep, xu2020attribution, kapishnikov2021guided, smilkov2017smoothgrad}.

\noindent \textbf{CAM methods} One popular form of saliency maps is class activation mapping (CAM) \cite{zhou2016learning}, which sums activations within a layer of the network to produce heatmaps, weighted by a value related to the output classification. We consider two variants of CAM known as GradCAM and GradCAM++. GradCAM weights using the average gradient with respect to the desired output classification \cite{selvarajuGradCAMVisualExplanations}, and GradCAM++ builds on this idea but uses second-order gradients to produce explanations with improved object localization \cite{chattopadhay2018grad}. 

\noindent \textbf{IG methods} On the other hand, Integrated Gradients (IG) linearly interpolates between a baseline input (in our case, a black image) and the target input while summing the gradient of the output along the path \cite{sundararajanAxiomaticAttributionDeep}. In a variant called BlurIG, the path is not linear but generated by constantly blurring the original image using the Laplacian of Gaussian kernels \cite{xu2020attribution}. Meanwhile, Guided IG follows an adaptive path along pixels with the smallest derivative with respect to the output \cite{kapishnikov2021guided}.

\noindent \textbf{Other methods} We also looked at two methods unrelated to IG and CAM. SmoothGrad averages the gradient of the classification output with respect to noisy version of the input image \cite{smilkov2017smoothgrad}. This method can also be combined with other methods such as IG, but we used the method with vanilla gradients highlighted in the paper. Meanwhile, LIME approximates the model behavior in the neighborhood of a given input using a simple linear model to generate sparse explanations \cite{ribeiroWhyShouldTrust2016}.

\subsection{Saliency Metrics}
Although defining which explanations are helpful or unhelpful can be a challenging task \cite{adebayoSanityChecksSaliency2018}, several qualitative characteristics for good explanations have been proposed, including fidelity to model prediction and generalizability across explanations \cite{sundararajanAxiomaticAttributionDeep, yehFidelitySensitivityExplanations2019, felHowGoodYour2022}. Various quantitative metrics have been developed to examine these properties, but we found these methods either required unnatural, out-of-distribution perturbations or have focused on examples that were not meaningful for explaining typical neural network use cases.

\noindent \textbf{Model perturbation.} Some methods randomize parts or all of the weights in a neural network and expect explanations to change \cite{adebayoSanityChecksSaliency2018, adebayoDebuggingTestsModel2020}, but we question whether the effects of manual changing sections of a network on corresponding explanations can be reliably predicted. Instead, we do this in a less artificial way by saving checkpoints of the model as it is being trained, ensuring we are able to produce models in the same way as a typical user might in the process of training of fine-tuning models. 

\noindent \textbf{Input perturbation.} Similarly, many metrics perturb inputs and observe how explanations change in order to measure the quality of an explanation \cite{yehFidelitySensitivityExplanations2019, tomsettSanityChecksSaliency2020, zhouFeatureAttributionMethods2022, kapishnikovXRAIBetterAttributions2019}. In all examples we investigated, these perturbed inputs are not in the training distribution, and we believe it is difficult to justify that explanations should change or stay the same. In contrast, we use augmentations which are in the training distribution to guarantee that the network should behave in the same way as in training and thus should explain predictions in the same way.

\noindent \textbf{Generating ground truth explanations.} Zhou \etal randomizes dataset labels to coincide solely with a single image augmentation, implying this augmentation is the ground-truth explanation for this dataset \cite{zhouFeatureAttributionMethods2022}. Similarly, the BAM dataset generates an artificial dataset by pasting images from one dataset to another and training models in such a way that the feature importance is known \cite{yang2019benchmarking}. Fel \etal bootstraps networks with different test sets and anticipates the explanations to be the same between networks which trained on a given input and those which only encounter the input in the test set \cite{felHowGoodYour2022}. While these methods are informative, they fail to capture the full story of a typical usage of neural networks on a natural dataset. 

\noindent \textbf{Human studies.} Zimmermann \etal tries to evaluate the usefulness of visual explanations by having users predict network activations with and without visual aids \cite{zimmermannHowWellFeature2021}. Using human studies is sensible given the explanations are meant to improve human understanding, but can be difficult to formulate and expensive to implement. Quantitative methods can help analyze other factors and narrow down methods to examine more closely \cite{zhouEvaluatingQualityMachine2021}.

\noindent \textbf{Similarity between explanations.} Methodology for determining similarity or distance scores between two explanations quantitatively is not evident a priori, and prior works are somewhat divided between various methods including Spearman rank correlation \cite{tomsettSanityChecksSaliency2020, felHowGoodYour2022}, structural similarity index (SSIM) \cite{adebayoDebuggingTestsModel2020, adebayoSanityChecksSaliency2018}, and Pearson correlation on the histogram of gradients for each explanation \cite{adebayoSanityChecksSaliency2018}. We chose in this work to use structural similarity index because of its applicability to images based on human perception. We explored using Pearson correlation instead of SSIM and found similar results presented in the supplementary material.

\section{Method}
\subsection{Problem Formulation}
We focus on evaluating the performance of saliency methods on supervised classification models trained on a set of data points $D = \{(x_1, y_1), (x_2, y_2), \ldots, (x_n, y_n)\}$ where $x \in X$ is an input image and $y \in Y$ is the class label of the image. A model learns a function $f:X\to Y$ that estimates $\hat{y}$ from the given $x$, where $\hat{y}$ is as close as possible to $y$. A saliency method $\Phi$ tries to estimate a map $M_i \in [0, 1]^{m\times n}$ from a given input image $x_i \in \mathbb{R}^{m\times n \times c}$, its corresponding output $\hat{y_i} \in \mathbb{R}^d$, and the model $f$. More formally, saliency maps can be represented as:

\begin{equation}
    \Phi(f, x_i) = M_i
    \label{eq:saliency-map}
\end{equation}

To evaluate saliency methods, we measure changes in $M_i$ by varying either $x_i$ or $f$. The next subsections discuss these modifications to the input and the model, and propose measurements on saliency maps that capture their performance and reliability.

\subsubsection{Data Augmentations \label{sec:method-data-aug}}
The data augmentation module applies \textit{natural} image transformations to represent image variation observed in the wild. To ensure these augmentations are simple, replicable, and reversible, we use a subset of transformations in TrivialWideAugment \cite{muller2021trivialaugment}, which randomly applies a single augmentation with random magnitude. Each transformation is either a fixed magnitude (e.g. flipping the image) or uniformly sampled from a discrete, linearly spaced set of 61 magnitudes. We removed transformations we deemed to be not naturally occurring, such as shearing. We define photometric transformations as those which vary the perceived colors of the images (e.g. varying the contrast), whereas geometric transformations vary the orientation of the images (e.g. rotation, translation). We classify geometric transformations as the set $G$ and photometric transformations as the set $H$ and let $T = G \cup H$.

A model $f$ is invariant to the data transformation $t\in T$ if $f(x_i) \equiv f(t(x_i))$. We define a saliency map $\Phi(f, x_i)$ as being equivalent to $\Phi(f, t(x_i))$ if it is equivariant to geometric data transformations and invariant to photometric data transformations. In other words, for $h\in H$, we expect $\Phi(f, x_i) = \Phi(f, h(x_i))$, while for $g\in G$, we reverse the operation on the saliency maps before evaluation, meaning we expect $\Phi(f, x_i) = g^{-1}(\Phi(f, g(x_i)))$. 
 
\begin{figure}
    \begin{center}
    \includegraphics[scale=0.35]{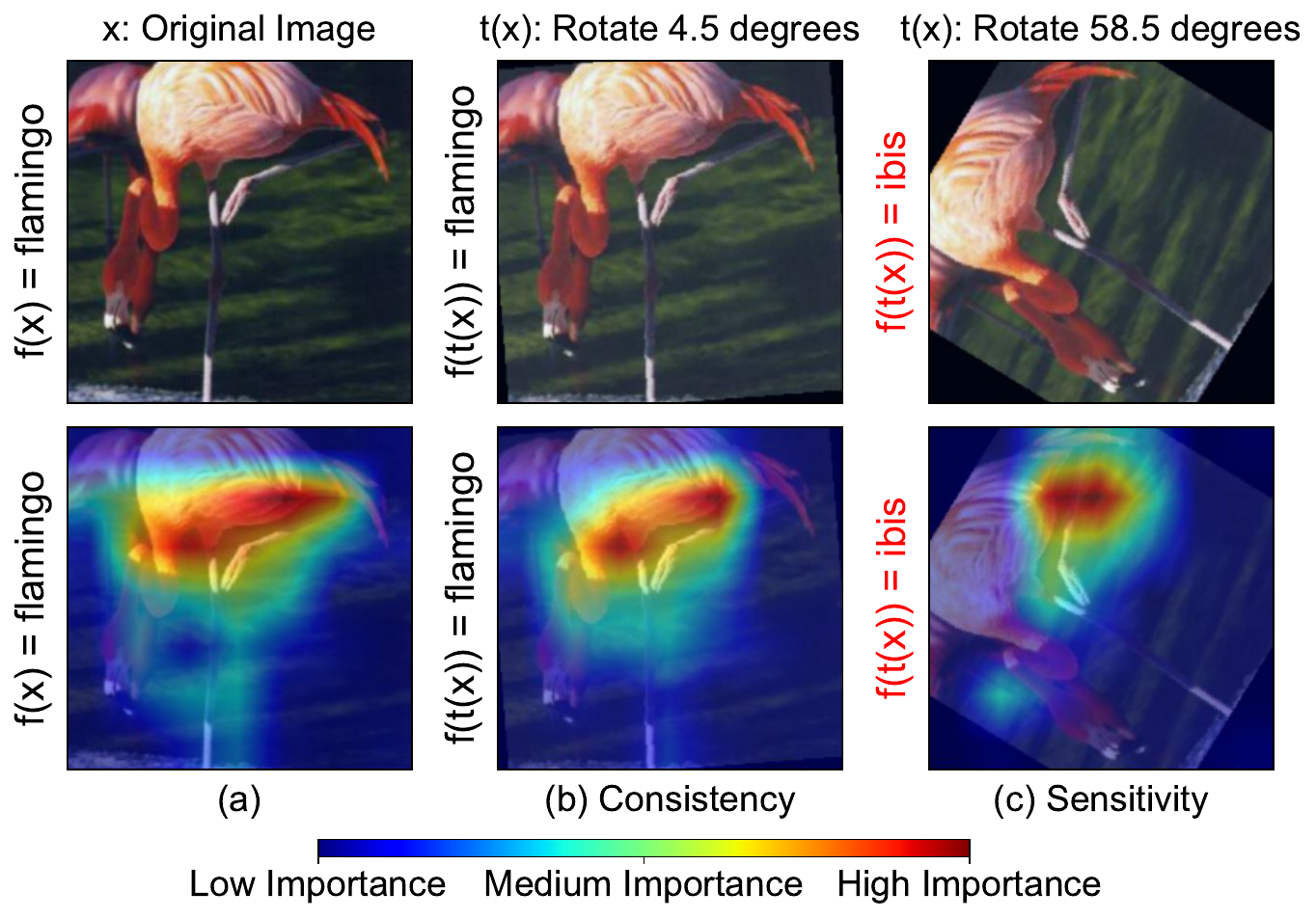}
    \end{center}
    \vspace{-4mm}
    \caption{\textbf{GradCAM consistency and sensitivity to data transformations on ResNet50/Caltech101}: the \textbf{top} row shows the input images, and the \textbf{bottom} row shows the corresponding saliency maps. In the reference image (a), the model correctly classifies the original image as a \textit{flamingo}. (b) displays GradCAM's \textbf{consistency} as the model correctly classifies the transformed image ($f(x) \equiv f(t(x_i))$) with a similar saliency map as (a). (c) displays GradCAM's \textbf{sensitivity} where the model incorrectly classified the transformed image as \textit{ibis} $\left( f(x) \not \equiv f(t(x)) \right)$ and the saliency map emphasizes differently from (a).}
    \label{fig:data-transform}
    \vspace{-2mm}
\end{figure}

 On the other hand, data transformations that result to a different model output $f(x_i) \not \equiv f(t(x_i))$ should also correspond to different saliency maps $\Phi(f, x_i) \not \equiv \Phi(f, t(x_i))$. Figure \ref{fig:data-transform} shows this on a sample image using GradCAM where consistent model behavior should result to equivalent saliency maps, and changing model behavior should result to different saliency maps.

\subsubsection{Model Augmentations \label{sec:method-model-aug}}
Prior to training, models have random weights and are unable to classify properly. As a model learns, the underlying weights change and adapt to the data presented. Given the main goal of saliency maps is to clarify the behavior of the underlying model, saliency maps should display the model changes as it undergoes training. When the model updates from $f' \rightarrow f$, the corresponding saliency map should also evolve $\Phi(f', x_i) \rightarrow \Phi(f, x_i)$.

To have realistic changes in model weights, we capture the changing model as it is trained from the first epoch until it reaches the final trained state. In the final state, the model should have learned where and how to \textit{look} at the images and classify images correctly. We quantify this performance using the test set classification accuracy. In other words, we should see that as a model learns, the saliency maps should reflect the increasing accuracy of this changing model.

\subsection{Proposed Metrics}
Structural Similarity Index Measure (SSIM) \cite{wang2004imagessim} is used on the saliency maps to quantify the deviation of maps due to variations from data and model augmentations. Equation \ref{eq:ssim} defines the similarity of two maps $M_x$ and $M_y$ using $\operatorname{SSIM}$, which lies between 0 and 1. The variable $\mu_{M_x}$ is the pixel sample mean of $M_x$, $\sigma_{M_x}^2$ is the variance of $M_x$, and $C_1 = 0.01, C_2 = 0.03$ are variables to stabilize the division for small denominator values. Subsequent sections on the proposed metrics will use this similarity measure for comparing two output saliency maps.
\begin{equation}
    \operatorname{SSIM}(M_x, M_y) = \frac{ \left(2\mu_{M_x}\mu_{M_y} + C_1 \right) \left(2\sigma_{M_x,M_y} + C_2 \right) } { \left(\mu_{M_x}^2 + \mu_{M_y}^2 + C_1 \right) \left(\sigma_{M_x}^2 + \sigma_{M_y}^2 + C_2 \right)}
    \label{eq:ssim}
\end{equation}
\subsubsection{Consistency}
Based on the idea described in \S~\ref{sec:method-data-aug}, we propose the \textbf{consistency} metric. The metric measures the robustness of saliency maps to data augmentations. Given a model robust to a set of data augmentations, reliable saliency maps should show equivalent explanations for the input $x_i$ and its transformed counterpart $t(x_i)$ (Equation \ref{eq:consistency1}). 
\begin{equation}
    f(x_i) \equiv f(t(x_i)) \implies 
    \Phi(f, x_i) \equiv \Phi(f, t(x_i))
    \label{eq:consistency1}
\end{equation}

Let $(X, H)^*$ be a set such that for $x\in X$ and $h\in H$, $(x, h) \in (X, H)^*$ if and only if $f(x) \equiv f(h(x))$ and similarly for $(X, G)^*$. We evaluate the robustness of a given method $\Phi$ based on the similarity of the two maps and propose the following consistency metric:
\begin{multline}
    \operatorname{consistency} = \frac{1}{N} \sum_{(x, h)\in (X, H)^*} \operatorname{SSIM}(\Phi(f, x_i), \Phi(f, h(x_i))) \\ + \frac{1}{N} \sum_{(x, g)\in (X, G)^*} \operatorname{SSIM}(\Phi(f, x_i), g^{-1}(\Phi(f, g(x_i)))),
    \label{eq:consistency}
\end{multline}
where $N = |(X, H)^* \cup (X, G)^*|$.
\subsubsection{Sensitivity}
Complementing the idea of consistency, if a model prediction changes due to either a change in input (\S~\ref{sec:method-data-aug}) or a change in the model itself (\S~\ref{sec:method-model-aug}), an optimal saliency method should also reflect these changes. We call this characteristic \textbf{sensitivity}. A saliency method should be sensitive to the underlying changes in the model itself (Equation \ref{eq:model-fidelity1}) or to the response of a model to an input augmentation $t \in T$ (Equation \ref{eq:img-fidelity1}).
\begin{equation}
    f(x_i) \not \equiv f'(x_i) \implies 
    \Phi(f, x_i) \not \equiv \Phi(f', x_i)
    \label{eq:model-fidelity1}
\end{equation}
\begin{equation}
    f(x_i) \not \equiv f(t(x_i)) \implies
    \Phi(f, x_i) \not \equiv \Phi(f, t(x_i))
    \label{eq:img-fidelity1}
\end{equation}

We reformulate minimizing $\operatorname{SSIM}$ to instead maximize $d(M_1, M_2) = 1-\operatorname{SSIM}(M_1, M_2)$ to maintain a similar notation as the consistency metric. Let $(x, h)\in (X, H)'$ for $x\in X$ and $h \in H$ if and only if $f(x) \not \equiv f(h(x))$ and similarly for $(X, G)'$. Furthermore, let $(x, f') \in (X, F)'$ for $x\in X$ if and only if $f(x) \not \equiv f'(x)$ for a naturally perturbed model $f'$. We propose the following sensitivity metric:
\begin{multline}
    \operatorname{sensitivity} = \frac{1}{M} \sum_{(x, h)\in (X, H)'} \operatorname{d}(\Phi(f, x_i), \Phi(f, h(x_i))) \\ + \frac{1}{M} \sum_{(x, g)\in (X, G)'} \operatorname{d}(\Phi(f, x_i), g^{-1}(\Phi(f, g(x_i)))) \\ + \frac{1}{M} \sum_{(x, f)\in (X, F)'} \operatorname{d}(\Phi(f, x_i), \Phi(f', x_i)),
    \label{eq:sensitivity}
\end{multline}
where $M = |(X, F)' \cup (X, G)' \cup (X, H)'|.$
\subsubsection{COSE}
Saliency methods should satisfy both consistency and sensitivity. Consistency enforces saliency methods to be robust to input changes that don't affect the model. Sensitivity imposes saliency methods to reflect changes that do affect the model. An optimal saliency map should balance between these two metrics. Thus, we combine these into a single metric \textbf{COnsistency-SEnsitivity (COSE)} using their harmonic mean. This allows evaluation by only looking at a single metric, enabling faster and easier estimation of saliency method performances. To achieve a high COSE, a method should have both high consistency and sensitivity.

\begin{equation}
    \operatorname{COSE} = \frac{2 \cdot \operatorname{sensitivity} \cdot \operatorname{consistency}}{\operatorname{sensitivity} + \operatorname{consistency}} \times 100\%
\end{equation}

\subsection{Evaluation Setup}

\noindent \textbf{Models.} We trained eight types of models with five datasets. The models are variations of four base models: ResNet50 \cite{heDeepResidualLearning2016}, ConvNext \cite{liu2022convnet}, ViT-B/16 \cite{dosovitskiy2020vit}, and Swin-T \cite{liu2021swin} to cover convolutional-based models and transformer-based models. Each model was trained to achieve at least 75\% average accuracy on the test set across all datasets. The settings and performances of all models trained on each dataset are provided in the appendix. Supervised and unsupervised training for each model was also considered. Models were pre-trained on ImageNet \cite{deng2009imagenet} using self-supervised learning methods DINO \cite{caron2021emerging}, MoCov3 \cite{chen2021empirical}, iBOT \cite{zhou2021ibot}, and SparK \cite{tian2023designing}, respectively. The models were then fine-tuned on the downstream task of image classification.

\noindent \textbf{Datasets.} The datasets CIFAR-10 \cite{krizhevsky2009learning}, Caltech 101 \cite{li_andreeto_ranzato_perona_2022}, Caltech-UCSD Birds (CUB200) \cite{welinder2010caltech}, EuroSAT \cite{helber2019eurosat}, and Oxford 102 flowers (Oxford102) \cite{nilsback2008automated} were used in the evaluation of saliency methods on classification tasks. These were chosen to look at the performance of saliency methods across a variety of data, ranging from fine-grained to coarse-grained datasets. 

\noindent \textbf{Saliency Methods.} The methods GradCAM \cite{selvarajuGradCAMVisualExplanations}, GradCAM++ \cite{chattopadhay2018grad}, IG \cite{sundararajanAxiomaticAttributionDeep}, BlurIG \cite{xu2020attribution}, Guided IG \cite{kapishnikov2021guided}, SmoothGrad \cite{smilkov2017smoothgrad}, and LIME \cite{ribeiroWhyShouldTrust2016} were analyzed in this paper. Each of these saliency methods were evaluated for all types of models and for all datasets. The recommended parameters from the corresponding papers of the saliency methods were used and are provided in the appendix.

\noindent \textbf{Data Transformations.} We apply two sets of data transformations for images: photometric and geometric. Photometric transformations involve changes in blur, contrast, brightness, equalization, smoothness, sharpness, and color. Geometric transformations consider translation, rotation, and flipping. These were applied during training to make sure the model is invariant to both types of transformations.
\section{Results and Analysis}
We present findings from running evaluations on different saliency methods, and their performances based on our proposed metrics consistency, sensitivity, and COSE.
\begin{figure}
    \begin{center}
    \includegraphics[scale=0.49]{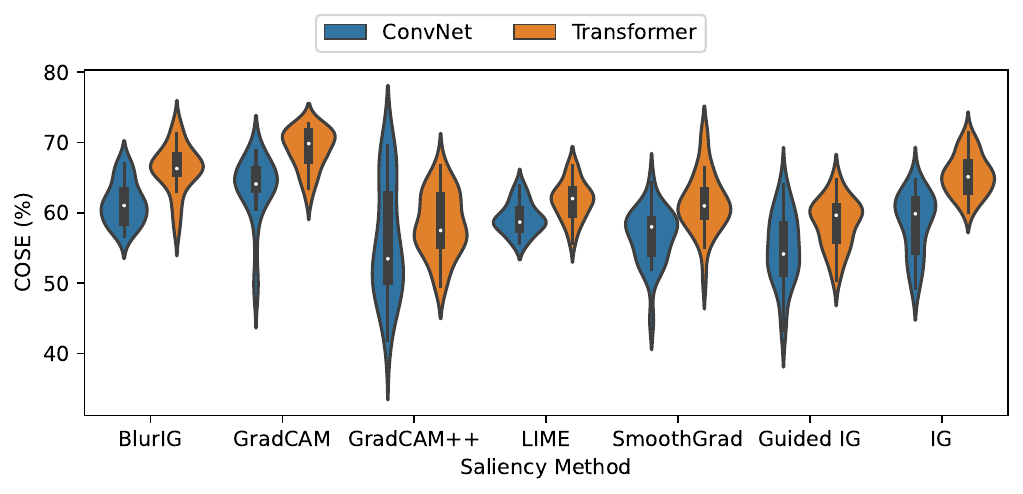}
    \end{center}
    \vspace{-4mm}
    \caption{\textbf{The distributions of COSE for ConvNets and Transformers compared for all saliency methods, shown as a violin plot.} Within each violin, the thin line shows the 1.5x interquartile range, the thick line shows the interquartile range, and the white dot shows the median. The shape of the violin shows how data points are distributed. Transformers outperform ConvNets on average for all methods.}
    \label{fig:transformerconvnet}
    \vspace{-2mm}
\end{figure}
\begin{figure*}
    \begin{center}
    \includegraphics[scale=0.36]{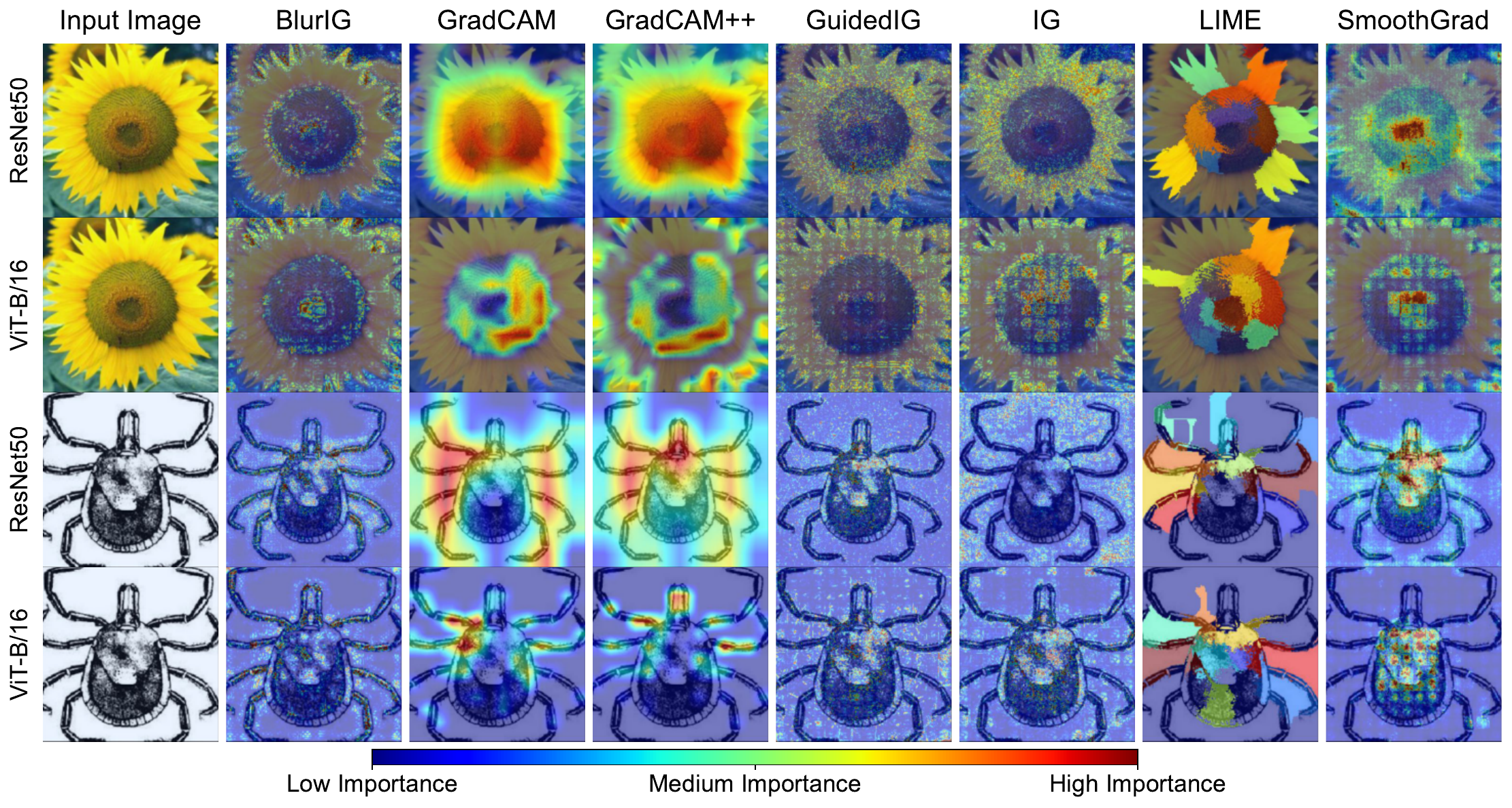}
    \end{center}
    \vspace{-5mm}
    \caption{\textbf{Results of saliency methods on ViT-B/16 (a transformer model), ResNet50 (a convolutional model), and Oxford102/Caltech101.} We qualitatively observe the explanations for ViT-B/16 to be similar (Guided IG, SmoothGrad) or better (BlurIG, IG, LIME, GradCAM, GradCAM++). In general, we find transformer model explanations are more coherent than convolutional model explanations.}
    \label{fig:saliency-images-sup-unsup}
    \vspace{-3mm}
\end{figure*}

\subsection{Transformers have better explanations}

\noindent \textbf{Transformer explanations had higher COSE and sensitivity for all methods.} Figure \ref{fig:transformerconvnet} shows transformer model explanations consistently outperforming those of ConvNets. Table \ref{tab:cose-saliency-methods} supports this even further, with transformers obtaining a higher average COSE score than ConvNets for every dataset and saliency method. Transformers also displayed higher sensitivity than ConvNets for almost every dataset and saliency method. In terms of consistency, although we observed ConvNets outperformed transformers for CAM-based methods and CUB, the difference is negligible when looking at the overall performance. Figure \ref{fig:saliency-images-sup-unsup} illustrates how explanations appear more coherent for ViT-B/16 than ResNet50.

\noindent \textbf{Transformer vs ConvNet receptive fields could explain the difference in saliency maps.} The self-attention mechanism of transformers doing patch-wise operations allow for better interpretability due to the availability of a global view of the image. ConvNets, on the other hand, use local operators that have limited receptive fields, restricting the amount of information that can be utilized by saliency methods \cite{luo2016understanding, he2021transreid}. In addition, given unsupervised vision transformers have been found to outperform similarly-trained ConvNets in terms of various segmentation tasks \cite{caron2021emerging}, we speculate this better spatial understanding may extend to explanations of vision transformers as well. We explore this further by looking at supervised and unsupervised network comparisons in the appendix.

\subsection{GradCAM is more reliable than other methods}

\noindent \textbf{GradCAM has the highest COSE for most of the experiments.} Table \ref{tab:cose-saliency-methods} shows the performance of different saliency methods across all datasets and models. In 65\% of the evaluation settings, GradCAM outperformed other saliency methods, with BlurIG having the highest COSE for 22.5\% of the experiments, IG for 5\%, GradCAM++ for 5\%, and GuidedIG for 2.5\% of the experiments. 
Although COSE is a descriptive single metric for overall performance, we also look at the performance of saliency methods on consistency and sensitivity individually to give further insight into what contributes to the performance of the saliency methods. 

\begin{table*}[!ht]
    \small
    \vskip 0.15in
    \begin{center}
    \begin{tabular}{l | l | c c c c c c c}
    \toprule
    \textbf{Dataset} & \textbf{Model} & \textbf{BlurIG} & \textbf{GradCAM} & \textbf{GradCAM++} & \textbf{GuidedIG} & \textbf{IG} & \textbf{LIME} & \textbf{SmoothGrad} \\
     &  & \cite{xu2020attribution} & \cite{selvarajuGradCAMVisualExplanations} & \cite{chattopadhay2018grad} & \cite{kapishnikov2021guided} & \cite{sundararajanAxiomaticAttributionDeep} & \cite{ribeiroWhyShouldTrust2016} & \cite{smilkov2017smoothgrad} \\
    \midrule 
    Caltech101 & ConvNext & 63.01\% & \underline{63.50\%} & \textbf{65.28\%} & 54.25\% & 62.53\% & 61.51\% & 60.90\% \\
     & ResNet50 & \textbf{65.77\%} & \underline{61.86\%} & 45.41\% & 52.41\% & 58.80\% & 56.69\% & 59.60\% \\
     & Swin-T & \textbf{67.81\%} & \underline{67.15\%} & 57.50\% & 52.54\% & 64.68\% & 63.17\% & 61.95\% \\
     & ViT-B/16 & \textbf{69.60\%} & \underline{68.66\%} & 60.30\% & 57.48\% & 66.67\% & 61.12\% & 66.41\% \\
    \midrule
    CIFAR10 & ConvNext & 61.46\% & \textbf{66.74\%} & \underline{66.72\%} & 53.02\% & 60.91\% & 62.06\% & 58.85\% \\
     & ResNet50 & \underline{60.47\%} & \textbf{62.29\%} & 42.75\% & 48.27\% & 51.20\% & 59.86\% & 52.02\% \\
     & Swin-T & \underline{66.35\%} & \textbf{69.66\%} & 58.29\% & 50.11\% & 63.05\% & 65.76\% & 56.95\% \\
     & ViT-B/16 & \underline{66.46\%} & \textbf{71.54\%} & 59.11\% & 55.72\% & 66.68\% & 63.85\% & 61.00\% \\
    \midrule
    CUB200 & ConvNext & 54.15\% & \underline{60.61\%} & 59.81\% & 59.59\% & \textbf{61.20\%} & 56.90\% & 47.89\% \\
     & ResNet50 & \underline{58.63\%} & 44.01\% & 40.74\% & 56.50\% & \textbf{60.05\%} & 55.50\% & 52.38\% \\
     & Swin-T & 62.37\% & \underline{62.51\%} & 49.78\% & 60.14\% & \textbf{64.02\%} & 59.06\% & 56.05\% \\
     & ViT-B/16 & 59.07\% & \textbf{64.80\%} & 56.65\% & \underline{61.26\%} & 60.42\% & 58.31\% & 53.70\% \\
    \midrule
    EuroSAT & ConvNext & 59.17\% & \textbf{65.47\%} & \underline{63.58\%} & 52.83\% & 61.45\% & 60.23\% & 57.17\% \\
     & ResNet50 & 57.28\% & \textbf{62.27\%} & 45.17\% & 40.87\% & 46.14\% & \underline{59.47\%} & 47.96\% \\
     & Swin-T & \underline{64.74\%} & \textbf{66.49\%} & 53.82\% & 47.51\% & 61.99\% & 59.60\% & 60.43\% \\
     & ViT-B/16 & 67.85\% & \textbf{70.31\%} & 57.95\% & 57.90\% & \underline{68.12\%} & 60.63\% & 62.48\% \\
    \midrule
    Oxford102 & ConvNext & 58.60\% & \textbf{62.78\%} & \underline{61.74\%} & 58.73\% & 60.71\% & 57.23\% & 57.77\% \\
     & ResNet50 & \underline{61.13\%} & \textbf{61.90\%} & 40.30\% & 54.93\% & 55.07\% & 58.32\% & 56.60\% \\
     & Swin-T &\underline{66.57\%} & \textbf{67.38\%} & 56.87\% & 52.86\% & 63.42\% & 62.06\% & 59.41\% \\
     & ViT-B/16 & \underline{66.90\%} & \textbf{68.31\%} & 59.14\% & 59.67\% & 65.92\% & 61.44\% & 62.95\% \\
    \midrule
     & Overall & \underline{63.23\%} & \textbf{64.66\%} & 54.59\% & 54.73\% & 61.33\% & 60.11\% & 57.94\% \\
    \bottomrule
    \end{tabular}
    \caption{\textbf{COSE score of different saliency methods on various models and datasets.} For each dataset and model combination, the best saliency method is \textbf{bolded} and the second-best is \underline{underlined}. For most models and datasets, GradCAM has the highest COSE, followed by BlurIG and IG. It can also be seen that most methods (GradCAM, GradCAM++, BlurIG) apart from IG and GuidedIG struggle with CUB200.}
    \label{tab:cose-saliency-methods}
    \end{center}
\end{table*}

\begin{figure*}
    \begin{center}
    \includegraphics[scale=0.65]{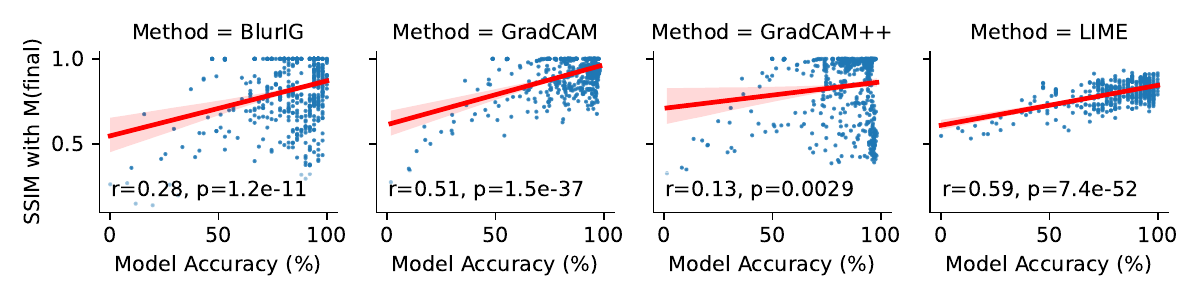}
    \includegraphics[scale=0.65]{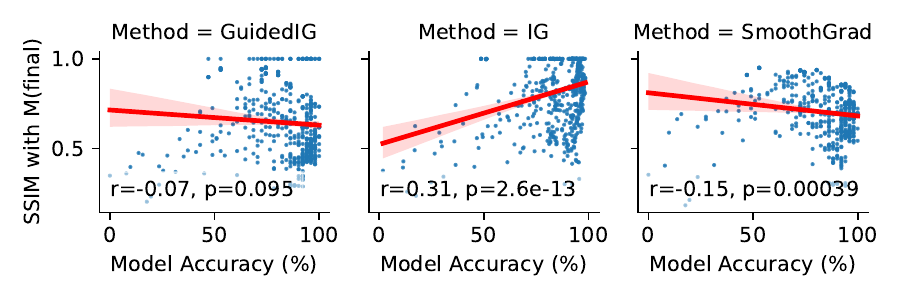}
    \end{center}
    \vspace{-0.4cm}
    \caption{\textbf{SSIM of the saliency map of the final trained model $M_{final}$ with respect to the saliency map of a partially trained model $M_e$}. GradCAM and LIME generally has increasing SSIM with increasing model accuracy. The correlation (r) and the corresponding p-values (p) are also annotated in each plot. We use alpha=0.05 (correlation for GuidedIG is insignificant).}
    \label{fig:ssim-vs-accs}
\end{figure*}

\noindent \textbf{GradCAM can reflect changing model behavior.} Figure \ref{fig:ssim-vs-accs} shows the relationship between the model accuracy at a collection of intermediate model training epochs $e$ and the difference in saliency maps $M_{final}$ and $M_e$ ($\operatorname{SSIM}(M_e, M_{final})$). $M_{final}$ is the saliency map for a fully trained model, and $M_e$ is the saliency map of an untrained or a partially trained model. Both LIME and GradCAM show a significant positive correlation between SSIM and accuracy, indicating that saliency maps from these methods can illustrate changes in model performance. 

\begin{figure}
    \begin{center}
    \includegraphics[scale=0.34]{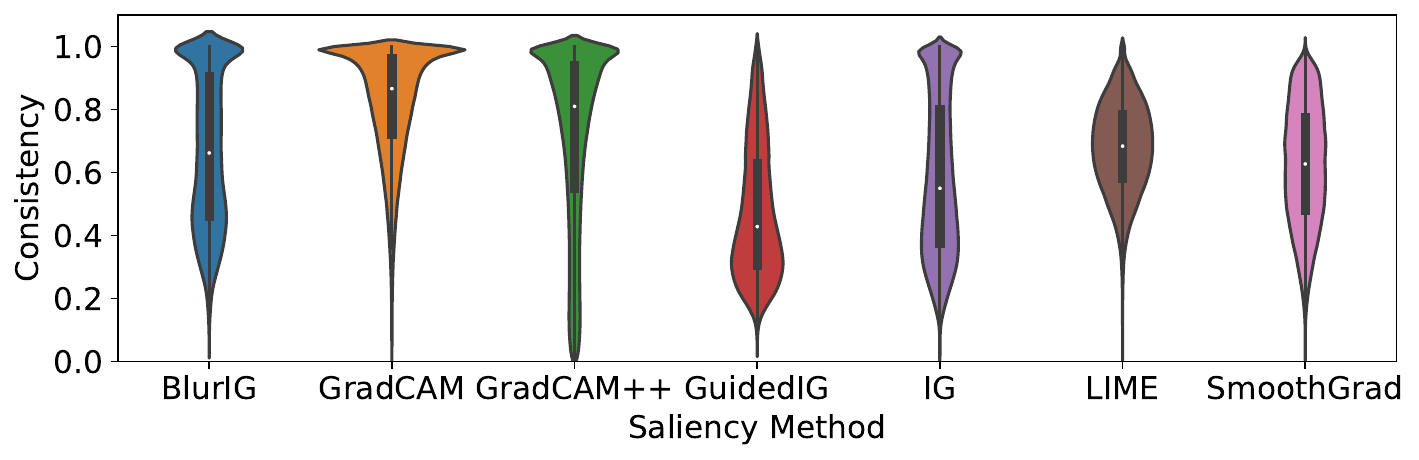}
    \end{center}
    \vspace{-0.4cm}
    \caption{\textbf{GradCAM has the highest average consistency. }We can also observe the distribution of consistency across different samples, with GradCAM having more samples with high consistency.}
    \label{fig:saliency-consistency}
    \vspace{-2mm}
\end{figure}

\noindent \textbf{GradCAM is more robust to data transformations.} Looking at the consistency metrics in Figure \ref{fig:saliency-consistency}, GradCAM has the highest average consistency. The general distribution also shows GradCAM having more samples with high consistency values when compared to other methods. This indicates that GradCAM, followed by GradCAM++ and BlurIG, are robust to  data transformations that do not affect model behavior.

\begin{figure}
    \begin{center}
    \includegraphics[scale=0.34]{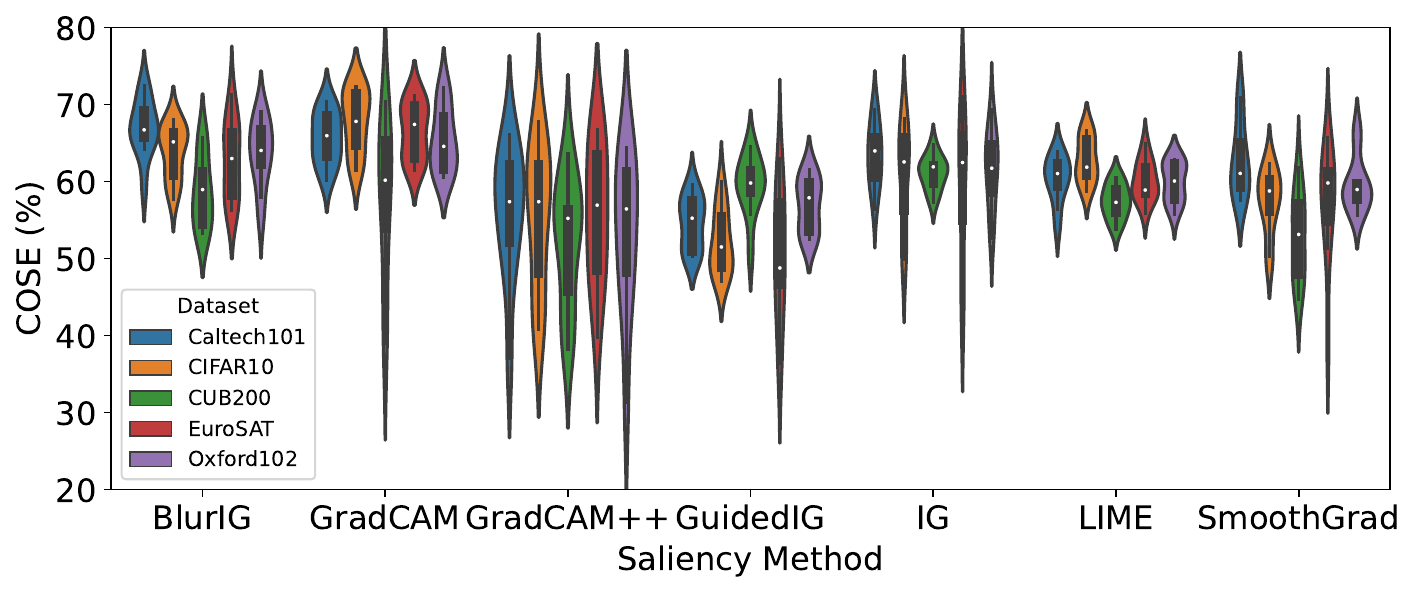}
    \end{center}
    \vspace{-4mm}
    \caption{\textbf{Most saliency methods struggle with CUB200.} Looking at each saliency method group, CUB200 has the lowest average performance for all methods except GuidedIG and IG. }
    \label{fig:saliency-dataset}
    \vspace{-2mm}
\end{figure}

\noindent \textbf{Limitations of GradCAM. }
Although GradCAM is shown to do better than other methods for most of the models and datasets, it evidently struggles with CUB200. Table \ref{tab:cose-saliency-methods} shows the COSE for various saliency methods across datasets and metrics. It also shows GradCAM has low scores on CUB200. Figure \ref{fig:saliency-dataset} shows that across different saliency methods, CUB200 has the lowest average COSE. This could be contributed to the tendency of CAM methods to emphasize larger areas of importance. Unlike GradCAM and GradCAM++, IG and GuidedIG focus on specific details (also see Figure \ref{fig:saliency-images} for sample results), and are observed to perform better on CUB200 based on COSE. The ability to distinguish between small differences on fine-grained datasets like CUB200 can significantly affect the performance of a saliency method. 

\subsection{How do we improve existing saliency methods? }
\noindent \textbf{Balancing consistency and sensitivity.} GradCAM is shown to outperform other saliency methods on several angles. However, with a COSE of 64.66\%, GradCAM still has areas for improvement. Figure \ref{fig:saliency-images} shows that GradCAM tends to predict general areas, which limits its sensitivity to model changes. Due to the large salient area presented, it's more difficult to isolate differences due to model changes. SmoothGrad and BlurIG show more specific areas, but they tend to be unstable to input perturbations. Future work on saliency methods should aim to balance performance on both - being robust while maintaining good sensitivity.

\begin{figure}
    \begin{center}
    \includegraphics[scale=0.55]{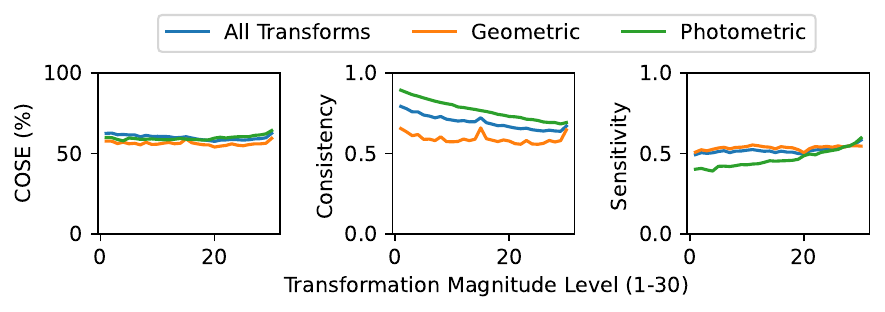}
    \end{center}
    \vspace{-4mm}
    \caption{\textbf{Average performance for all methods on all transformations and separately for geometric and photometric transformations.} While photometric transformations have decreasing sensitivity and increasing fidelity as transformation magnitudes increase, geometric transformations seem to have approximately the same performance regardless of transformation magnitude.}
    \label{fig:transformvsmag}
    \vspace{-2mm}
\end{figure}

\noindent \textbf{Methods should consider both geometric and photometric consistency.} Figure \ref{fig:transformvsmag} shows saliency methods generally have lower consistency and higher sensitivity as transformation magnitudes increase, but ultimately average to a stable COSE over all transformation magnitudes. Splitting into geometric and photometric transformations, we observe in the same figure that this trend is mostly for photometric transformations, as saliency methods perform about the same even when geometric transformations increase. This suggests that saliency methods struggle in different ways for photometric and geometric transformations. While COSE gives an overview of overall performance across all transform magnitudes, we recommend saliency method developers consider photometric changes and geometric changes as separate problems while trying to achieve consistency in both.

\section{Conclusion}

We presented an evaluation pipeline measuring two crucial characteristics for saliency methods - consistency, which requires images with the same classification to have the same explanation, and sensitivity, which describes that images with different classifications to have different explanations. We combine these two measures into a single metric COSE which is only maximized by balancing the two properties. By applying natural augmentations to images in arbitrary datasets, we show our metrics can emphasize the advantages and the limitations of saliency methods when ground truth model explanations are not available.

Through our metrics, we analyzed the performance of seven commonly used saliency methods across five datasets and eight models. Fundamentally, our metric COSE is best-suited for saliency metrics whose explanations closely reflect the prediction of the network - giving similar explanations for the consistent model behavior and contrasting explanations for different model behavior. We see our work as a starting point for researchers to further explore and improve saliency methods for better model understanding.

\paragraph{Acknowledgements.} The project was funded in part by NSF grant \#1749833 to Subhransu Maji. The experiments were performed on the University of Massachusetts GPU cluster funded by the Mass. Technology Collaborative.

{\small
\bibliographystyle{ieee_fullname}
\bibliography{egbib}
}

\end{document}